%% file: main.tex
\documentclass[runningheads]{llncs}

\PassOptionsToPackage{numbers, compress}{natbib}

\usepackage[final]{AdvML_Frontiers_2024}

\usepackage[utf8]{inputenc} % allow utf-8 input
\usepackage[T1]{fontenc}    % use 8-bit T1 fonts
\usepackage{url}            % simple URL typesetting
\usepackage{booktabs}       % professional-quality tables
\usepackage{amsfonts}       % blackboard math symbols
\usepackage{nicefrac}       % compact symbols for 1/2, etc.
\usepackage{microtype}      % microtypography
\usepackage{xcolor}         % colors

\usepackage{subfig}
\usepackage{float}
\usepackage{wrapfig}

\usepackage{graphicx}
\usepackage{booktabs}
\usepackage{multirow}

\usepackage[accsupp]{axessibility}  % Improves PDF readability for those with disabilities.

\usepackage{algorithm}
\usepackage{algpseudocode}
\usepackage{amsmath,bm}
\usepackage{amsfonts}

\usepackage[pagebackref,breaklinks,colorlinks]{hyperref}
\usepackage{hyperref}

\usepackage{orcidlink}

\title{Unveiling Synthetic Faces: \\How Synthetic Datasets Can Expose Real Identities}

\author{%
Hatef Otroshi Shahreza$^{1,2}$ and S\'{e}bastien Marcel$^{1,3}$ \\\vspace{-1pt}
$^{1}$Idiap Research Institute, Martigny, Switzerland\\
$^{2}$\'{E}cole Polytechnique F\'{e}d\'{e}rale de Lausanne (EPFL), Lausanne, Switzerland\\
$^{3}$Universit\'{e} de Lausanne (UNIL), Lausanne, Switzerland \\
\texttt{\{hatef.otroshi,sebastien.marcel\}@idiap.ch} \\
}

\begin{document}

\maketitle

\begin{figure}[htbp]
    \centering
    \begin{minipage}{0.05\textwidth}
        \rotatebox{90}{Real}
    \end{minipage}%
    \begin{minipage}{0.93\textwidth}
        % First row of subfigures
        \subfloat{\includegraphics[width=0.16\linewidth]{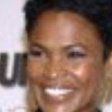}} \hfill
        \subfloat{\includegraphics[width=0.16\linewidth]{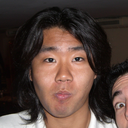}} \hfill
        \subfloat{\includegraphics[width=0.16\linewidth]{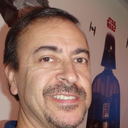}} \hfill
        \subfloat{\includegraphics[width=0.16\linewidth]{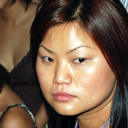}} \hfill
        \subfloat{\includegraphics[width=0.16\linewidth]{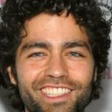}} \hfill
        \subfloat{\includegraphics[width=0.16\linewidth]{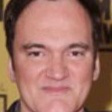}} \hfill
        \\\vspace{-9.5pt}
    \end{minipage}

    \begin{minipage}{0.05\textwidth}
        \rotatebox{90}{\hspace{10pt}Synthetic}
    \end{minipage}%
    \begin{minipage}{0.93\textwidth}
        \captionsetup[subfloat]{labelformat=empty}
        % Second row of subfigures
        \subfloat[DCFace]{\includegraphics[width=0.16\linewidth]{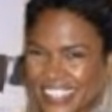}} \hfill
        \subfloat[\scalebox{0.9}{IDiff-Face\scalebox{0.9}{ (U)}}]{\includegraphics[width=0.16\linewidth]{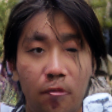}} \hfill
        \subfloat[\scalebox{0.9}{IDiff-Face\scalebox{0.9}{ (T)}}]{\includegraphics[width=0.16\linewidth]{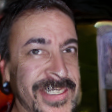}} \hfill
        \subfloat[\scalebox{0.9}{GANDiffFace}]{\includegraphics[width=0.16\linewidth]{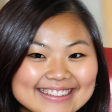}} \hfill
        \subfloat[IDNet]{\includegraphics[width=0.16\linewidth]{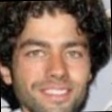}} \hfill
        \subfloat[SFace]{\includegraphics[width=0.16\linewidth]{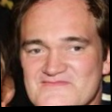}} \hfill
    \end{minipage}
    \caption{Sample face images leaked from training data (first row) of generative models in different state-of-the-art synthetic face recognition datasets (second row).}
    \label{fig:sample_firstpage}
\end{figure}

\begin{abstract}
Synthetic data generation is gaining increasing popularity in different computer vision applications. Existing state-of-the-art face recognition models are trained using large-scale face datasets, which are crawled from the Internet and raise privacy and ethical concerns. To address such concerns, several works have proposed generating synthetic face datasets to train face recognition models. However, these methods depend on generative models, which are trained on real face images. In this work, we design a simple yet effective membership inference attack to systematically study if any of the existing synthetic face recognition datasets leak any information from the real data used to train the generator model. We provide an extensive study on 6 state-of-the-art synthetic face recognition datasets, and show that in all these synthetic datasets, several samples from the original real dataset are leaked. To our knowledge, this paper is the first work which shows the leakage from training data of generator models into the generated synthetic face recognition datasets. Our study demonstrates privacy pitfalls in synthetic face recognition datasets and paves the way for future studies on generating responsible synthetic face datasets. Project page: \href{https://www.idiap.ch/paper/unveiling_synthetic_faces/}{https://www.idiap.ch/paper/unveiling\_synthetic\_faces}
\end{abstract}

\section{Introduction}\label{sec:intro}
\vspace{-10pt}

Recent advancements in state-of-the-art face recognition models are achieved by training deep neural networks with penalty-based softmax loss functions on large-scale datasets \cite{deng2019arcface,kim2022adaface,george2024edgeface,meng2021magface}.
Existing face recognition datasets contain millions of face images, such as VGGFace2 \cite{cao2018vggface2}, MS-Celeb-1M \cite{guo2016ms}, WebFace260M \cite{zhu2021webface260m}, which are typically collected by crawling the Internet without proper user consent. This raises ethical and legal concerns about the use of such datasets for the development of face recognition models. 
In particular, recent data regulation frameworks, such as the European Union Artificial Intelligence Act, further support the rights of subjects whose data are used in a dataset to train such models.  
Consequently, several face recognition datasets, including but not limited to \cite{cao2018vggface2,guo2016ms}, have been retracted by their creators, to prevent potential legal issues. 
Therefore, the availability of such large-scale datasets and the possibility of research in the face recognition task has become uncertain.

Recently, generating synthetic face recognition datasets has  emerged as a promising alternative to large-scale real datasets and has become a promising solution to address the  ethical and legal 
concerns \cite{boutros2022sface,kolf2023identity,kim2023dcface,boutros2023idiff,melzi2023gandiffface,bae2023digiface,shahreza2024sdfr,synthdistill-ijcb}. 
The generation of synthetic face datasets, however, relies on the development of face generative models, which enable the generation of synthetic samples from the probability distribution of a real face dataset.
Meanwhile, most synthetic face recognition datasets are built upon Generative Adversarial Networks (GANs) \cite{boutros2022sface,kolf2023identity} or Diffusion Models (DMs) \cite{kim2023dcface,boutros2023idiff}.
These face generative models are trained on a dataset of real face images, and therefore leakage of information from the training dataset to the generated face images can raise privacy concerns in the generated synthetic face recognition datasets. 
Along the same lines, several studies in the literature of generative models have shown the memorization issue in different models~\cite{carlini2023extracting,vyas2023provable,somepalli2023diffusion,somepalli2023understanding,li2024shake,carlini2021extracting,nasr2023scalable,zhang2022membership,tinsley2021face}, which sparks concerns on the application of generative models for privacy-sensitive problems. 
In this paper, we design a simple yet effective membership inference attack against synthetic face datasets to systematically study if any of the existing samples in the generated dataset leak any information from the real data used to train the generator model. 
We provide an extensive evaluation of 6 state-of-the-art synthetic face recognition datasets, and demonstrate that in all these synthetic datasets, several samples from original real datasets are leaked.
Fig.~\ref{fig:sample_firstpage} illustrates sample face images from 6 state-of-the-art synthetic face recognition datasets that are leaked from the training set of their generator models.
To our knowledge, this paper presents the first work which shows the leakage from training data of generator models into synthetic face recognition datasets.
Our study demonstrates privacy pitfalls in synthetic face recognition datasets and paves the way for future studies on generating responsible synthetic face datasets.

In the remainder of the paper, we first review state-of-the-art synthetic face recognition datasets in the literature in Section~\ref{sec:related-work}. Then, we describe our membership inference attack and present our evaluation of state-of-the-art synthetic face recognition datasets in Section~\ref{sec:membership-inference-attack}. In Section~\ref{sec:discussion}, we discuss our findings and explain the limitations of our study as well as current shortcomings in the literature. Finally, the paper is concluded in Section~\ref{sec:conclusion}.

\vspace{-5pt}
\section{Related Work}\label{sec:related-work}
\vspace{-10pt}
As discussed earlier, existing synthetic face recognition datasets are typically generated using a generative model. While some papers used pretrained face generative models (such as pretrained StyleGAN on the FFHQ dataset), other works retrained a generative model on another dataset or proposed a new face generator model.
Boutros et al. \cite{boutros2022sface} used the StyleGAN2-ADA  \cite{StyleGAN2} as their generative model and trained it on the CASIA-WebFace dataset \cite{yi2014learning} with identities serving as class labels. Then, they utilized their identity-conditioned StyleGAN2-ADA model to generate the SFace dataset of generated face images, and demonstrated its effectiveness for training face recognition algorithms.

Kolf et al. \cite{kolf2023identity} also used StyleGAN2-ADA  \cite{StyleGAN2} trained on the CASIA-WebFace dataset \cite{yi2014learning} and proposed a three-player GAN framework to generate the IDNet dataset. Their three-player framework integrates identity information into the image generation of StyleGAN, where the third player is used to force the generator network to generate identity-separable face images.

In contrast to most works in the literature that used GAN-based generator models, some works generated synthetic datasets using diffusion models. Kim et al. \cite{kim2023dcface},  introduced the dual (identity and style) condition face generator based on a diffusion model and trained it on the CASIA-WebFace \cite{yi2014learning} dataset. They used a patch-wise style extractor combined with a time-step dependent ID loss to train their generator model. Then, they generated the DCFace dataset by synthesizing different identities using identity condition and also different samples per identity using the style condition. 
They published two versions of their dataset, DCFace-0.5M and DCFace-1.2M, where the smaller version is a subset of the larger one.

In \cite{boutros2023idiff}, the authors trained a latent diffusion model conditioned on identity features obtained from a pretrained face recognition model using the FFHQ dataset.  
For sample generation, they used their trained diffusion model to generate different identities by randomly sampling the identity context from a uniform distribution. 
In another approach, they used an unconditional diffusion model to generate different identities (two-stage). To generate different samples per identity, they fixed the identity condition and changed the latent noise. Considering these two different generation approaches, they proposed two datasets called IDiff-Face (Uniform) and IDiff-Face (Two-stage), respectively.

In \cite{melzi2023gandiffface}, Melzi et al. introduced the GANDiffFace dataset, which is generated using both GAN-based and diffusion-based generators. They first used a pretrained StyleGAN3~\cite{StyleGAN3} model (trained on FFHQ) to generate different identities, and then used a pretrained DreamBooth~\cite{ruiz2023dreambooth} model to generate different samples for each identity. DreamBooth~\cite{ruiz2023dreambooth} is a diffusion model based on the Stable Diffusion~\cite{rombach2022high} model that is trained on the LAION dataset~\cite{schuhmann2022laion}.

\begin{table}[tb]
	\centering
		\renewcommand{\arraystretch}{1.05}
		\setlength{\tabcolsep}{4.5pt}
		\caption{Synthetic Face Recognition Datasets in the Literature.}
		\scalebox{0.95}{
			\begin{tabular}{cllccc}
				\textbf{Reference}  &  \textbf{Synthetic Dataset} &  \textbf{Generator}  &  \textbf{Training Dataset}\\ 
				\toprule
				\multirow{1}{*}{\cite{boutros2022sface}}  & SFace & StyleGAN-ADA (identity-conditioned)  & CASIA-WebFace \\ \hline
				\multirow{1}{*}{\cite{kolf2023identity}}  & IDNet & StyleGAN-ADA (identity-conditioned)  & CASIA-WebFace \\  \hline
				\multirow{2}{*}{\cite{kim2023dcface}}  & \multirow{2}{*}{DCFace} & new diffusion model   & \multirow{2}{*}{CASIA-WebFace} \\  
                    & & (identity and style conditioned) \\ \hline
                    \multirow{2}{*}{\cite{boutros2023idiff}}  & IDiff-Face (Uniform) & new diffusion model   & \multirow{2}{*}{FFHQ} \\ 
                    &  IDiff-Face (Two-stage) & (identity-conditioned) \\\hline
                \multirow{2}{*}{\cite{melzi2023gandiffface}}  & \multirow{2}{*}{GANDiffFace} & StyleGAN (pretrained)  & FFHQ \\  
                & &  DreamBooth (pretrained) & LAION\\

				\bottomrule
			\end{tabular}\label{tab:related-works}
		} 
  \vspace{-2em}
\end{table}

Table \ref{tab:related-works} summarizes different synthetic face recognition datasets in the literature which are generated using generative models. We should note that another category of methods to generate synthetic datasets is computer-graphic-based methods, e.g., DigiFace-1M \cite{bae2023digiface}, which is excluded in this study.

\vspace{-5pt}
\section{Membership Inference Attack}\label{sec:membership-inference-attack}
\vspace{-10pt}

% \begin{figure}[tbh]
\begin{wrapfigure}{r}{0.61\textwidth}
\vspace{-2em}
    \centering
    \includegraphics[width=1\linewidth, trim={1cm 0cm 2cm 0.2cm},clip]{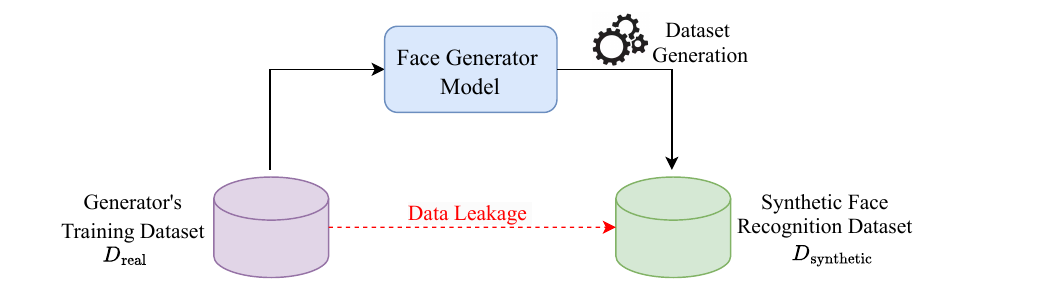}
    \caption{Schematic diagram of data leakage from generator's training data into generated synthetic face recognition dataset.}
    \label{fig:blockdiagram}
% \end{figure}
\vspace{-2em}
\end{wrapfigure}
As described in Section~\ref{sec:related-work}, synthetic face recognition datasets are often generated using a generator model. Let 
$D_\text{real}$ denote the real face recognition dataset that is used to train the generator model. Then, based on the synthetic dataset generation approach, the generator model is used to generate a synthetic face recognition dataset $D_\text{synthetic}$, which includes various synthetic identities and different sample images per each identity. Therefore, an important question is whether any of the generated images in the generated synthetic face recognition dataset $D_\text{synthetic}$ contain important information from training dataset  $D_\text{real}$, which was used to train the face generator model in the first place? In other words, do we have any leakage of information from $D_\text{real}$  to $D_\text{synthetic}$ ? Fig.~\ref{fig:blockdiagram} illustrates the process of leakage of information from the training dataset into the generated synthetic dataset.

We consider an exhaustive search approach to compare all possible pairs of images from $D_\text{real}$  and $D_\text{synthetic}$. To this end, we use an off-the-shelf face recognition model $F(.)$ to extract face embeddings from each face image, and then compare the embeddings of every pair of images from $D_\text{real}$  and $D_\text{synthetic}$. Then, we sort the pairs of images according to the similarity of embeddings and consider the top-$k$ pairs for visual comparison of images. Algorithm~\ref{algorithm} presents the pseudo-code of our approach.

\begin{figure}[t]
\vspace{-1.5em}
\begin{algorithm}[H]
	\caption{Membership Inference Attack against Synthetic Datasets (MIS).}\label{algorithm}
	\begin{algorithmic}[1]
		\Require: $D_\text{real}$, dataset used to train generator network; $D_\text{synthetic}$, generated synthetic face dataset; $F$, an off-the-shelf face recognition model; SIM, similarity function (e.g., cosine similarity) to compare two embeddings extracted by face recognition $F$; $k$, number of top similar pairs to return for visual comparison.
		\Procedure{MIS Attack}{}%{Training}{params}
		\State Initialize list $\mathcal{S}=[\hspace{3pt}]$
		\For{$I_{\text{synthetic},i}\in D_\text{synthetic}$}
		\For{$I_{\text{real},j}\in D_\text{real}$}
		\State $s= \text{SIM}( F(I_{\text{synthetic},i}),  F(I_{\text{real},j}) )$
		\State $\mathcal{S}$.append(s)
		\EndFor
		\EndFor
            \State return $\mathcal{S}$.sort$()[0:k]$
		\EndProcedure
	\end{algorithmic}
\end{algorithm}
\vspace{-5em}
\end{figure}

For our experiments, we use an off-the-shelf face recognition model with ResNet100 backbone which is trained with AdaFace~\cite{kim2022adaface} loss function on the WebFace12M dataset~\cite{zhu2021webface260m}. We use the method presented in algorithm \ref{algorithm} to find pairs with high cosine similarity scores from the training dataset of the generator model and the generated synthetic dataset for all the synthetic datasets in Table \ref{tab:related-works}. Figs.~\ref{fig:sample_dcface}-\ref{fig:sample_sface} of Appendix illustrate sample face images from the training dataset of generator models which are leaked in the generated synthetic dataset for 6 state-of-the-art synthetic datasets, including   DCFace~\cite{kim2023dcface}, IDiff-Face (Uniform)~\cite{boutros2023idiff}, IDiff-Face (Two-stage)~\cite{boutros2023idiff},  GANDiffFace~\cite{melzi2023gandiffface},  IDNet~\cite{kolf2023identity}, and SFace~\cite{boutros2022sface} datasets, respectively.
The corresponding training dataset of generator models used to generate each synthetic dataset is reported in Table~\ref{tab:related-works}.
In the case of GANDiffFace~\cite{melzi2023gandiffface} which uses two pretrained generator models, we compare with the training dataset of StyleGAN which was used in the first stage to generate different synthetic identities. For the evaluation of the DCFace dataset, we consider its smaller set (i.e., DCFace 0.5M), which is also included in the larger version of this dataset.
As Figs.~\ref{fig:sample_dcface}-\ref{fig:sample_sface} of Appendix show, the generated synthetic datasets contain very similar images from the training set of their generator model, which raises concerns regarding the generation of such identities.

\input{hist}

We set the parameter $k$ in algorithm~\ref{algorithm} equal to 1500, and compared the retrieved pairs of images. However, in some cases, such as DCFace, we can easily find leaked matches in top-100 pairs.  
Fig. \ref{fig:histograms} illustrates the histogram of similarity scores of all retrieved pairs of images for each synthetic dataset and their corresponding values of similarity for top-k pairs (as dashed vertical lines). In addition, this figure depicts the similarity scores for positive and negative pairs in the IARPA Janus Benchmark–C (IJB-C)~\cite{maze2018iarpa} as a benchmark dataset using the same face recognition model used in our experiments. This figure also shows the threshold for False Accept Rate (FAR) of 0.01\% on IJB-C dataset (dotted vertical lines). Surprisingly, as can be seen Fig.~\ref{fig:histograms}, almost all retrieved pairs of  images in each synthetic dataset has a similarity greater than the matching threshold on IJB-C (FAR=0.01\%). That means if we have a face recognition system configured at this threshold, almost all retrieved pairs  of images (i.e., each image in the synthetic dataset and its closest image in the training dataset) are considered as the same identity. In other words, almost for each image in the generated synthetic dataset there is one image in the training dataset for which the face recognition model (configured at FAR=0.01\% on IJB-C) will recognise as the same identity. However, concluding the leakage based on the histogram of similarity scores has some limitations which are discussed in detail in Section~\ref{sec:discussion}.

We should note that the source code of our experiments as well as the file names of all images reported as leaked samples in this paper are publicly available in our  \href{https://www.idiap.ch/paper/unveiling_synthetic_faces/}{project page}\footnote{Project page: \href{https://www.idiap.ch/paper/unveiling_synthetic_faces/}{https://www.idiap.ch/paper/unveiling\_synthetic\_faces}} to facilitate the reproducibility of our study and help future researchers to build upon our findings.

\vspace{-5pt}
\section{Discussion}\label{sec:discussion}
\vspace{-10pt}
Our experiments 
demonstrate that state-of-the-art synthetic face recognition datasets contain samples that are very close to samples in the training data of their generator models.
In some cases the synthetic samples contain small changes to the original image, however, we can also observe in some cases the generated sample contains more variation (e.g., different pose, light condition, etc.) while the identity is preserved.
This suggests that the generator models are learning and memorizing the identity-related information from the training data and may generate similar identities. This creates critical concerns regarding the application of synthetic data in privacy-sensitive tasks, such as biometrics and face recognition.

The findings in our paper open several new research questions and introduce new research directions that require attention from the community:
\begin{itemize}
    \item In this paper, we used an exhaustive search approach to find samples which contain information leaked in the generated dataset. While our approach is effective in finding samples in the synthetic dataset that are similar to the training dataset, it requires comparing all possible pairs of images. However, comparing all possible pairs may not be efficient and such a membership inference attack can be deployed more efficiently. In particular, if the training data or generated data have larger samples\footnote{such as LAION~\cite{schuhmann2022laion} dataset used to train Stable Diffusion~\cite{rombach2022high} and includes 5 billion data.}, the required computation for all comparisons similarly increases.

    \item While our attack algorithm can find samples in the training data which are leaked into the synthetic dataset, it also returns several samples that are not necessarily obvious to contain leakage. 
    As an example, we can refer to three categories which we observed in our experiments: 
    \begin{enumerate}
        \item In some cases, the similar images found by our approach are images of children. However, not only the face recognition models also have a high error for children, but also distinguishing if two images are for the same child is difficult for human observers. Therefore, considering children's images for indicating a possible leakage is not reliable. This particularly happens for synthetic datasets whose generator models are trained on the FFHQ dataset, in which the population of young children is considerably large~\cite{or2020lifespan}. Fig.~\ref{fig:issues:children} of Appendix illustrates some samples of young people in IDiff-Face (Uniform),  IDiff-Face (Two-stage), and GANDiffFace datasets. We ignored such samples in our evaluations and did not recognise them as leaked samples.
        \item In some samples that were returned by our algorithm with high similarity scores, the synthetic image did not include a face image. In fact, these are unexpected samples in a face recognition dataset. Therefore, we ignored such samples in our evaluations. Fig.~\ref{fig:issues:no-face} of Appendix illustrates some examples of images that do not have face images from  DCFace,  IDNet, and SFace datasets.
        \item In some cases, the images with high similarity scores were not recognized as the same identity or were not convincing enough to a human observer to demonstrate leakage. Therefore, we ignored such samples in our evaluations. Fig.~\ref{fig:issues:no-match} of Appendix illustrates some examples of such images in which we could not conclude leakage in visual comparison for different datasets.
    \end{enumerate}
    
    As a result of having such samples in the output of our analyses, we needed to have a visual comparison step. While we found visual comparison necessary to draw valid conclusions, it requires a human observer, even for a small number of selected samples. In addition to the required human effort, it may introduce subjective bias in visual comparisons. We should note that the samples shown in Fig.~\ref{fig:sample_firstpage} and Figs.~\ref{fig:sample_dcface}-\ref{fig:sample_dcface_more} of Appendix were selected based on the unanimous agreement of several observers, however, such a human evaluation may not be efficient and consistent for future studies.

    \item An important future direction is to propose new measures to \textit{quantify} and benchmark the leakage of information from training datasets of generator models into the synthetic data. We would like to stress that proposing such a measure is not trivial and requires further research to address the previous question and eliminate the necessity of human observers. In particular, three  cases which we mentioned in previous point of our discussion (i.e., children images, no face images, not having same identity)  makes  statistical analyses (based on the similarity scores of retrieved pairs) a challenging task, and thus it is difficult to extract a reliable statistical metric for the leakage in the synthetic datasets using similarity scores. For example, histograms of scores for matched pairs from synthetic dataset and corresponding closest images from the training datasets in Fig.~\ref{fig:histograms} show that for almost all images in synthetic datasets, there is an image in the training dataset which is recognised as the same identity by a face recognition model. However, sample images in Figs.~\ref{fig:issues:children}-\ref{fig:issues:no-match} of Appendix, which are even among top-k retrieved pairs with high similarity score, are not necessarily recognised as a same identity by  a human observer, and therefore such samples cannot demonstrate identity leakage. Hence, evaluating the leakage based on similarity scores of retrieved pairs of images is not straightforward and requires further studies.

    \item Our experiments show that state-of-the-art synthetic datasets leak sensitive information from the training dataset of their generator models. Therefore, an important future direction is to generate responsible synthetic face datasets. This objective can be achieved by preventing such leakage in the data generation process or by further post-processing to eliminate such leakages. We should note that some of the existing datasets, such as DCFace~\cite{kim2023dcface}, have already tried to prevent such leakage in data generation. In \cite[Section 3.3]{kim2023dcface}, the authors explained that they removed samples that are more similar to images of training data than a predefined threshold. However, as images in Fig.~\ref{fig:sample_dcface} and Fig.~\ref{fig:sample_dcface_more} of Appendix show, such data cleaning has not been sufficient to prevent identity leakage in the generated synthetic dataset. This suggests that more efforts should be taken to avoid leakage in the synthetic data.
\end{itemize}

To draw the discussion to a close, we would like to highlight that the main motivation for generating synthetic datasets is to address privacy concerns in using large-scale web-crawled face datasets. Therefore, the leakage of any sensitive information (such as identities of real images in the training data) in the synthetic dataset spikes critical concerns regarding the application of synthetic data for  privacy-sensitive tasks, such as biometrics. Our study sheds light on the privacy pitfalls in the generation of synthetic face recognition datasets and paves the way for future studies toward generating responsible synthetic face datasets.

\vspace{-3pt}
\section{Conclusion}\label{sec:conclusion}
\vspace{-10pt}
In this paper, we explored the crucial question of \textit{``whether synthetic datasets expose real identities used for training their generator models?"}. We used a simple yet effective membership inference attack (based on exhaustive search) against synthetic datasets, and explored if any of the generated samples in the synthetic face recognition dataset leaks any information from the training dataset of the corresponding generator model. We evaluated 6 state-of-the-art synthetic face recognition datasets generated with different deep generative models (GAN-based and diffusion-based). We reported several samples for each dataset which demonstrate leakage of information in synthetic datasets from the training data of generator models. In some cases, the retrieved samples further indicate memorization or learning identity-related information in the generator models. We also discussed the limitations of our evaluation and outlined potential future directions. To our knowledge, this paper is first  work which shows the leakage from training data of generator models into synthetic face recognition dataset and reveals privacy pitfalls in the generation of synthetic face recognition datasets.

\section*{Acknowledgments}
\vspace{-10pt}
This research is based upon work supported by the Hasler foundation through the ``Responsible Face Recognition" (SAFER)  project as well as the H2020 TReSPAsS-ETN Marie Sk\l{}odowska-Curie early training network (grant agreement 860813).

% ---- Bibliography ----
\bibliographystyle{unsrtnat}
\bibliography{main}

\newpage
\appendix

\section{Sample Leaked Images}

Figs.~\ref{fig:sample_dcface}-\ref{fig:sample_sface} illustrate sample face images from the training dataset of generator models which are leaked in the generated synthetic dataset for 6 state-of-the-art synthetic datasets, including   DCFace~\cite{kim2023dcface}, IDiff-Face (Uniform)~\cite{boutros2023idiff}, IDiff-Face (Two-stage)~\cite{boutros2023idiff},  GANDiffFace~\cite{melzi2023gandiffface},  IDNet~\cite{kolf2023identity}, and SFace~\cite{boutros2022sface} datasets, respectively.
The corresponding training dataset of generator models used to generate each synthetic dataset is reported in Table~\ref{tab:related-works} of the paper. For, GANDiffFace~\cite{melzi2023gandiffface} which uses two pretrained generator models, we compared with the training dataset of StyleGAN which was used in the first stage to generate different synthetic identities.
As these figures show, the generated synthetic datasets contain very similar images from the training set of their generator model, which raises concerns regarding the generation of such identities. 
In some cases, such as in DCFace (Fig.~\ref{fig:sample_dcface} of Appendix), the similarity is very high, and the generated image has some small visual changes compared to the original training data. However, in some other datasets, the difference is higher, nevertheless, the identities of generated images look similar. 
We should note that images shown in Fig.~\ref{fig:sample_firstpage} and Figs.~\ref{fig:sample_dcface}-\ref{fig:sample_sface} of Appendix are some samples for each dataset, and for some of these synthetic datasets we can easily find more samples. For example, Fig.~\ref{fig:sample_dcface_more} of Appendix illustrates more samples in the DCFace dataset. 

\vspace{-1em}

\input{sample_leaked}

\newpage

\section{Sample of Difficult Matching Face Images}
\begin{figure}[H]
    \centering
    \begin{minipage}{0.05\textwidth}
        \rotatebox{90}{Real}
    \end{minipage}%
    \begin{minipage}{0.93\textwidth}
        % First row of subfigures
        \subfloat{\includegraphics[width=0.16\linewidth]{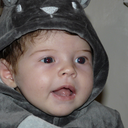}} \hfill
        \subfloat{\includegraphics[width=0.16\linewidth]{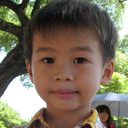}} \hfill
        \subfloat{\includegraphics[width=0.16\linewidth]{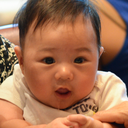}} \hfill
        \subfloat{\includegraphics[width=0.16\linewidth]{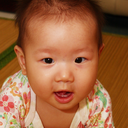}} \hfill
        \subfloat{\includegraphics[width=0.16\linewidth]{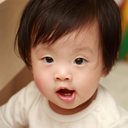}} \hfill
        \subfloat{\includegraphics[width=0.16\linewidth]{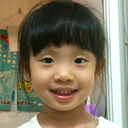}} \hfill
        \\\vspace{-9.5pt}
    \end{minipage}

    \begin{minipage}{0.05\textwidth}
        \rotatebox{90}{\hspace{10pt}Synthetic}
    \end{minipage}%
    \begin{minipage}{0.93\textwidth}
        \captionsetup[subfloat]{labelformat=empty}
        % Second row of subfigures
        \subfloat[\scalebox{0.9}{IDiff-Face\scalebox{0.9}{ (U)}}]{\includegraphics[width=0.16\linewidth]{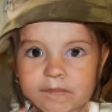}} \hfill
        \subfloat[\scalebox{0.9}{IDiff-Face\scalebox{0.9}{ (U)}}]{\includegraphics[width=0.16\linewidth]{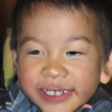}} \hfill
        \subfloat[\scalebox{0.9}{IDiff-Face\scalebox{0.9}{ (T)}}]{\includegraphics[width=0.16\linewidth]{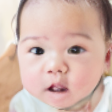}} \hfill
        \subfloat[\scalebox{0.9}{IDiff-Face\scalebox{0.9}{ (T)}}]{\includegraphics[width=0.16\linewidth]{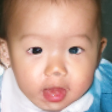}} \hfill
        \subfloat[\scalebox{0.9}{GANDiffFace}]{\includegraphics[width=0.16\linewidth]{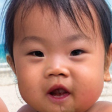}} \hfill
        \subfloat[\scalebox{0.9}{GANDiffFace}]{\includegraphics[width=0.16\linewidth]{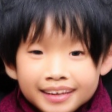}} \hfill
    \end{minipage}
    \caption{Sample face images of \textbf{children} which have high similarity, but we \textbf{ignored} them in our evaluations (i.e., we did not recognise them as leaked samples).}
    \label{fig:issues:children}
    \vspace{-2em}
\end{figure}

\begin{figure}[H]
    \centering
    
    \begin{minipage}{0.05\textwidth}
        \rotatebox{90}{\hspace{10pt}Synthetic}
    \end{minipage}%
    \begin{minipage}{0.93\textwidth}
        \captionsetup[subfloat]{labelformat=empty}
        % Second row of subfigures
        \subfloat[DCFace]{\includegraphics[width=0.16\linewidth]{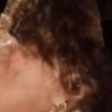}} \hfill
        \subfloat[DCFace]{\includegraphics[width=0.16\linewidth]{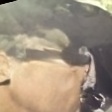}} \hfill
        \subfloat[IDNet]{\includegraphics[width=0.16\linewidth]{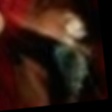}} \hfill
        \subfloat[IDNet]{\includegraphics[width=0.16\linewidth]{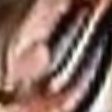}} \hfill
        \subfloat[SFace]{\includegraphics[width=0.16\linewidth]{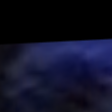}} \hfill
        \subfloat[SFace]{\includegraphics[width=0.16\linewidth]{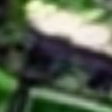}} \hfill
    \end{minipage}
    \caption{Sample  images which \textbf{do not include a face}, and we \textbf{ignored} them in our evaluations (i.e., we did not recognise them as leaked samples).}
    \label{fig:issues:no-face}
    \vspace{-2em}
\end{figure}

\begin{figure}[H]
    \centering
    \begin{minipage}{0.05\textwidth}
        \rotatebox{90}{Real}
    \end{minipage}%
    \begin{minipage}{0.93\textwidth}
        % First row of subfigures
        \subfloat{\includegraphics[width=0.16\linewidth]{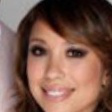}} \hfill
        \subfloat{\includegraphics[width=0.16\linewidth]{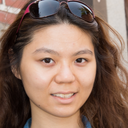}} \hfill
        \subfloat{\includegraphics[width=0.16\linewidth]{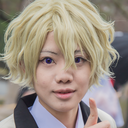}} \hfill
        \subfloat{\includegraphics[width=0.16\linewidth]{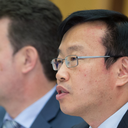}} \hfill
        \subfloat{\includegraphics[width=0.16\linewidth]{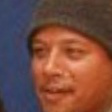}} \hfill
        \subfloat{\includegraphics[width=0.16\linewidth]{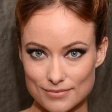}} \hfill
        \\\vspace{-9.5pt}
    \end{minipage}

    \begin{minipage}{0.05\textwidth}
        \rotatebox{90}{\hspace{10pt}Synthetic}
    \end{minipage}%
    \begin{minipage}{0.93\textwidth}
        \captionsetup[subfloat]{labelformat=empty}
        % Second row of subfigures
        \subfloat[DCFace]{\includegraphics[width=0.16\linewidth]{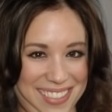}} \hfill
        \subfloat[\scalebox{0.9}{IDiff-Face\scalebox{0.9}{ (U)}}]{\includegraphics[width=0.16\linewidth]{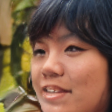}} \hfill
        \subfloat[\scalebox{0.9}{IDiff-Face\scalebox{0.9}{ (T)}}]{\includegraphics[width=0.16\linewidth]{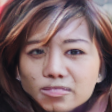}} \hfill
        \subfloat[\scalebox{0.9}{GANDiffFace}]{\includegraphics[width=0.16\linewidth]{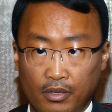}} \hfill
        \subfloat[IDNet]{\includegraphics[width=0.16\linewidth]{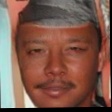}} \hfill
        \subfloat[SFace]{\includegraphics[width=0.16\linewidth]{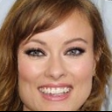}} \hfill
    \end{minipage}
    \caption{Sample face images of which have high similarity scores, but are not convincing for human observer to demonstrate data leakage in the synthesized image. Therefore, we \textbf{ignored} them in our evaluations (i.e., we did not recognise them as leaked samples).}
    \label{fig:issues:no-match}
    \vspace{1em}
\end{figure}

As discussed in Section~\ref{sec:discussion} of the paper, while our attack algorithm can find samples in the training data which are leaked into the synthetic dataset, it also returns several samples that are not necessarily obvious to contain leakage. Figs.~\ref{fig:issues:children}-\ref{fig:issues:no-match} illustrates sample of face images, which were difficult to match while having high  similarity score. We ignored such samples in our evaluation and did not recognise them as leaked samples.

\end{document}

%% file: hist.tex
% \begin{figure}[tbp]
\begin{wrapfigure}{r}{0.55\textwidth}
\vspace{-3.55em}
    \centering
    
        % \subfloat{\includegraphics[width=0.85\linewidth]{images/hists/IJB-C.pdf}}\vspace{-0.125cm} \\
        % \subfloat{\includegraphics[width=0.85\linewidth]{images/hists/DCFace_0_5m.pdf}}\vspace{-0.125cm} \\
        % \subfloat{\includegraphics[width=0.85\linewidth]{images/hists/IDiff-Face-uniform.pdf}}\vspace{-0.125cm} \\
        % \subfloat{\includegraphics[width=0.85\linewidth]{images/hists/IDiff-Face-two-stage.pdf}}\vspace{-0.125cm} \\
        % \subfloat{\includegraphics[width=0.85\linewidth]{images/hists/GANDiffFace.pdf}}\vspace{-0.125cm} \\
        % \subfloat{\includegraphics[width=0.85\linewidth]{images/hists/IDNet.pdf}}\vspace{-0.125cm} \\
        % \subfloat{\includegraphics[width=0.85\linewidth]{images/hists/SFace.pdf}}\vspace{-0.125cm} \\

\begin{minipage}{0.035\textwidth}
    \rotatebox{90}{\scalebox{0.825}{Benchmark}} % General vertical text for the remaining histograms
\end{minipage}%
\begin{minipage}{0.95\textwidth}
    \begin{minipage}{0.025\textwidth}
        \rotatebox{90}{\scalebox{0.60}{IJB-C}} % Vertical text for histogram 2
    \end{minipage}%
    \begin{minipage}{0.95\textwidth}
        \subfloat{\includegraphics[width=0.55\linewidth]{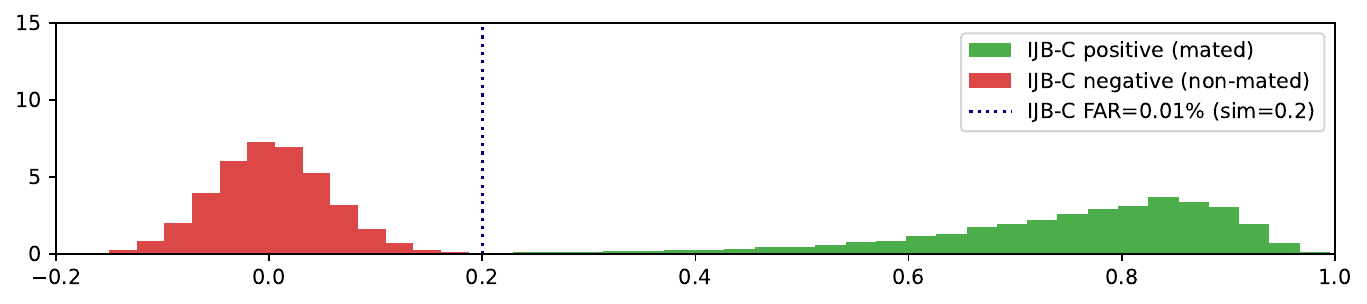}}\vspace{-0.125cm} \\
    \end{minipage}%
\end{minipage}%
\vspace{-0.20cm}
\rule{0.55\textwidth}{0.15mm} % Horizontal line separator
\vspace{-0.30cm}

\begin{minipage}{0.035\textwidth}
    \rotatebox{90}{\scalebox{0.825}{Synthetic Datasets}} % General vertical text for the remaining histograms
\end{minipage}%
\begin{minipage}{0.95\textwidth}
    \begin{minipage}{0.025\textwidth}
        \rotatebox{90}{\scalebox{0.60}{DCFace}} % Vertical text for histogram 2
    \end{minipage}%
    \begin{minipage}{0.95\textwidth}
        \subfloat{\includegraphics[width=0.55\linewidth]{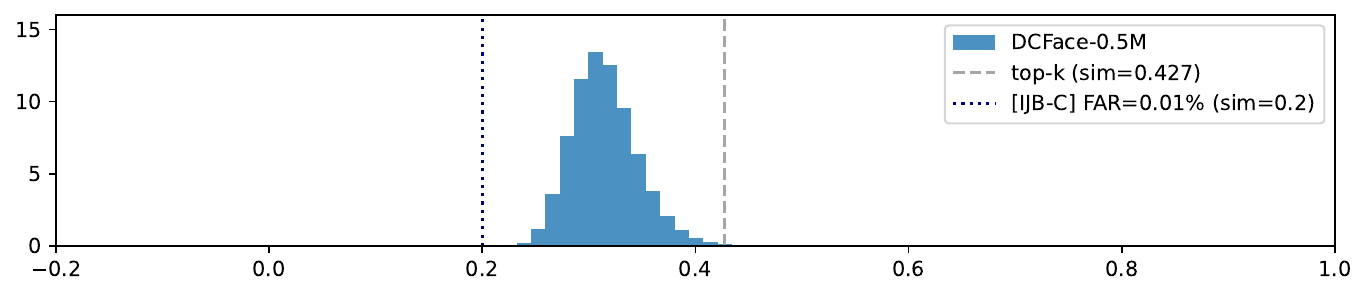}}\vspace{-0.125cm}
    \end{minipage}%
     \\
    \begin{minipage}{0.025\textwidth}
        \rotatebox{90}{\scalebox{0.60}{IDiff-Face (U)}} % Vertical text for histogram 2
    \end{minipage}%
    \begin{minipage}{0.95\textwidth}
        \subfloat{\includegraphics[width=0.55\linewidth]{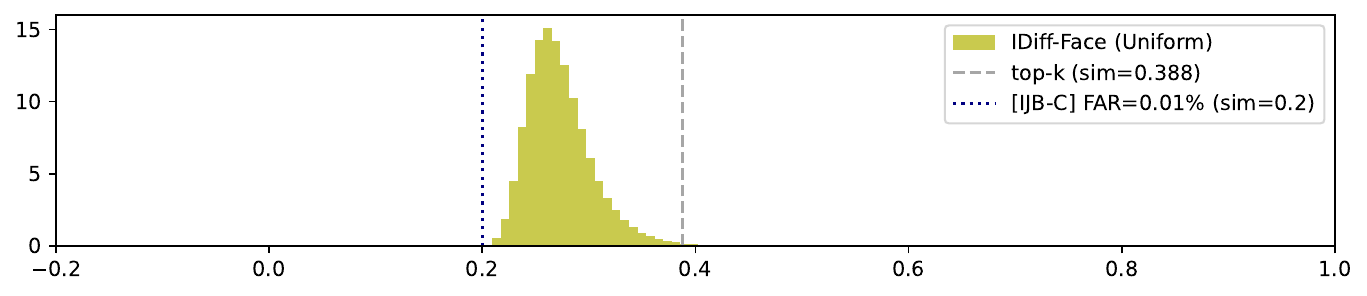}}\vspace{-0.125cm}
    \end{minipage}%
     \\
    \begin{minipage}{0.025\textwidth}
        \rotatebox{90}{\scalebox{0.60}{IDiff-Face (T)}} % Vertical text for histogram 2
    \end{minipage}%
    \begin{minipage}{0.95\textwidth}
        \subfloat{\includegraphics[width=0.55\linewidth]{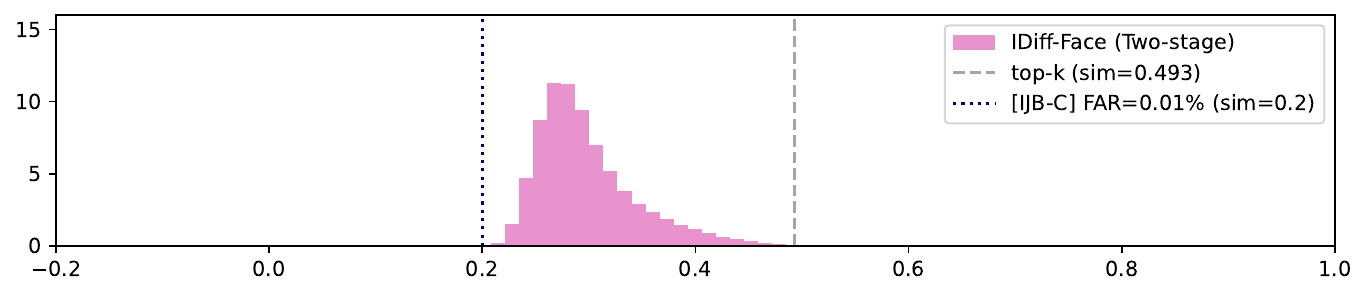}}\vspace{-0.125cm} 
    \end{minipage}%
    \\
    \begin{minipage}{0.025\textwidth}
        \rotatebox{90}{\scalebox{0.60}{GANDiffFace}} % Vertical text for histogram 2
    \end{minipage}%
    \begin{minipage}{0.95\textwidth}
        \subfloat{\includegraphics[width=0.55\linewidth]{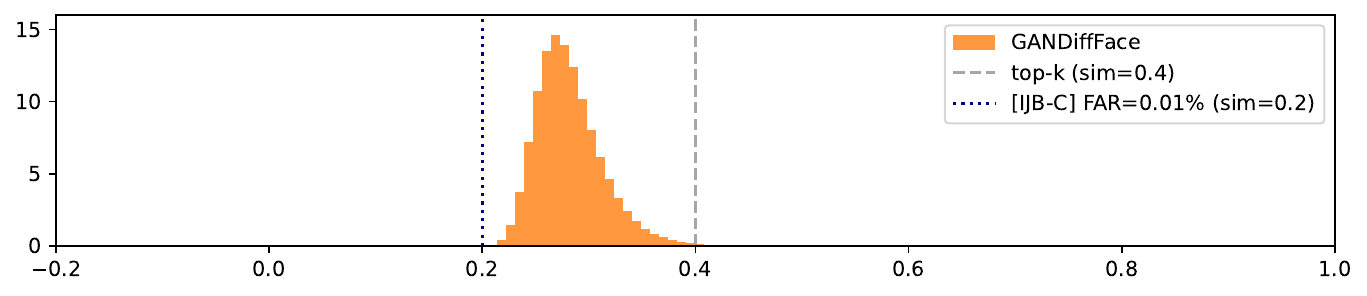}}\vspace{-0.125cm} 
    \end{minipage}%
    \\
    \begin{minipage}{0.025\textwidth}
        \rotatebox{90}{\scalebox{0.60}{IDNet}} % Vertical text for histogram 2
    \end{minipage}%
    \begin{minipage}{0.95\textwidth}
        \subfloat{\includegraphics[width=0.55\linewidth]{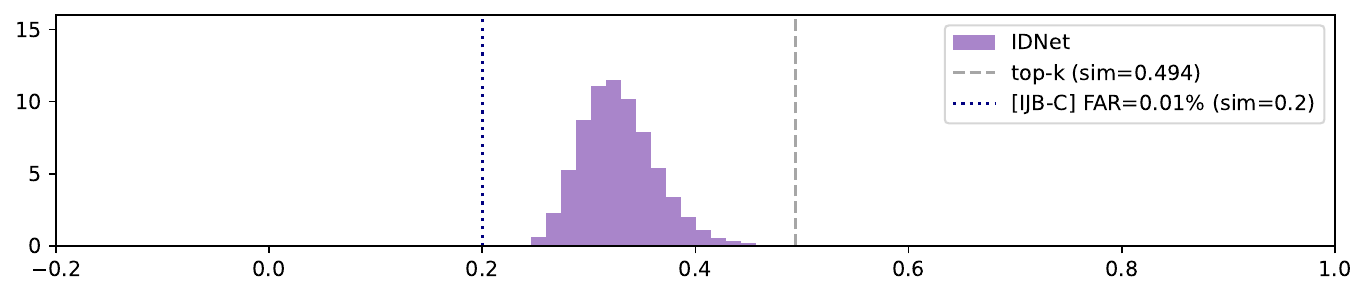}}\vspace{-0.125cm} 
    \end{minipage}%
    \\
    \begin{minipage}{0.025\textwidth}
        \rotatebox{90}{\scalebox{0.60}{SFace}} % Vertical text for histogram 2
    \end{minipage}%
    \begin{minipage}{0.95\textwidth}
        \subfloat{\includegraphics[width=0.55\linewidth]{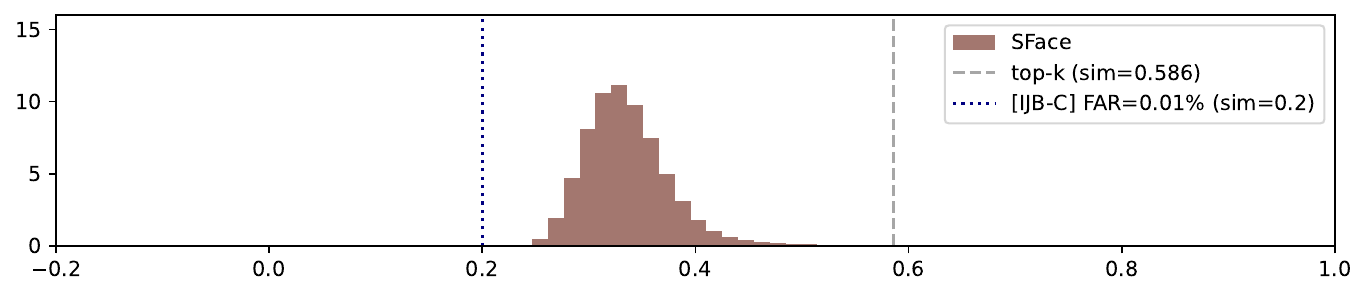}}\vspace{-0.125cm}
    \end{minipage}%
\end{minipage}%
\vspace{-0.025cm}
    \caption{Histogram of cosine similarity scores of all retrieved pairs of images for each synthetic dataset and their corresponding values of similarity for top-k pairs as dashed vertical lines (k=1500). The first plot shows the histogram of similarity scores for positive and negative pairs in IJB-C  dataset (benchmark) and the threshold for FAR=0.01\% on IJB-C with dotted vertical lines.}
    \label{fig:histograms}
    \vspace{-3em}
% \end{figure}
\end{wrapfigure}

%% file: sample_leaked.tex
\begin{figure}[htbp]
% \vspace{-2em}
    \centering
    \begin{minipage}{0.05\textwidth}
        \rotatebox{90}{Real}
    \end{minipage}%
    \begin{minipage}{0.93\textwidth}
        % First row of subfigures
        \subfloat{\includegraphics[width=0.16\linewidth]{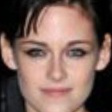}} \hfill
        \subfloat{\includegraphics[width=0.16\linewidth]{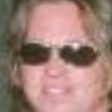}} \hfill
        \subfloat{\includegraphics[width=0.16\linewidth]{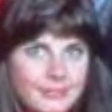}} \hfill
        \subfloat{\includegraphics[width=0.16\linewidth]{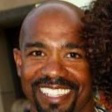}} \hfill
        \subfloat{\includegraphics[width=0.16\linewidth]{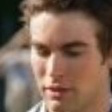}} \hfill
        \subfloat{\includegraphics[width=0.16\linewidth]{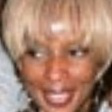}} \hfill
        \\\vspace{-9.5pt}
    \end{minipage}

    \begin{minipage}{0.05\textwidth}
        \rotatebox{90}{\hspace{5pt}Synthetic}
    \end{minipage}%
    \begin{minipage}{0.93\textwidth}
        % Second row of subfigures
        \subfloat{\includegraphics[width=0.16\linewidth]{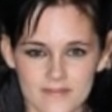}} \hfill
        \subfloat{\includegraphics[width=0.16\linewidth]{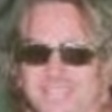}} \hfill
        \subfloat{\includegraphics[width=0.16\linewidth]{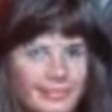}} \hfill
        \subfloat{\includegraphics[width=0.16\linewidth]{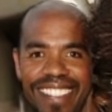}} \hfill
        \subfloat{\includegraphics[width=0.16\linewidth]{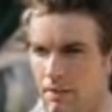}} \hfill
        \subfloat{\includegraphics[width=0.16\linewidth]{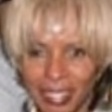}} \hfill
    \end{minipage}
    \vspace{-0.5em}
    \caption{Sample face images leaked from training data (first row) of the generative model in the \textbf{DCFace} dataset (second row). For more samples see Fig.~\ref{fig:sample_dcface_more}}
    \label{fig:sample_dcface}
    \vspace{-4em}
\end{figure}

\begin{figure}[htbp]
    \centering
    \begin{minipage}{0.05\textwidth}
        \rotatebox{90}{Real}
    \end{minipage}%
    \begin{minipage}{0.93\textwidth}
        % First row of subfigures
        \subfloat{\includegraphics[width=0.16\linewidth]{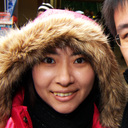}} \hfill
        \subfloat{\includegraphics[width=0.16\linewidth]{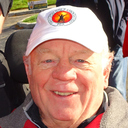}} \hfill
        \subfloat{\includegraphics[width=0.16\linewidth]{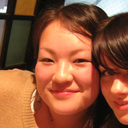}} \hfill
        \subfloat{\includegraphics[width=0.16\linewidth]{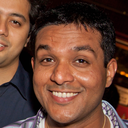}} \hfill
        \subfloat{\includegraphics[width=0.16\linewidth]{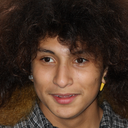}} \hfill
        \subfloat{\includegraphics[width=0.16\linewidth]{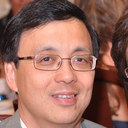}} \hfill
        \\\vspace{-9.5pt}
    \end{minipage}

    \begin{minipage}{0.05\textwidth}
        \rotatebox{90}{\hspace{5pt}Synthetic}
    \end{minipage}%
    \begin{minipage}{0.93\textwidth}
        % Second row of subfigures
        \subfloat{\includegraphics[width=0.16\linewidth]{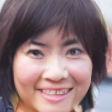}} \hfill
        \subfloat{\includegraphics[width=0.16\linewidth]{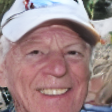}} \hfill
        \subfloat{\includegraphics[width=0.16\linewidth]{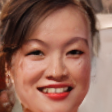}} \hfill
        \subfloat{\includegraphics[width=0.16\linewidth]{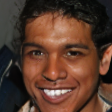}} \hfill
        \subfloat{\includegraphics[width=0.16\linewidth]{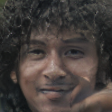}} \hfill
        \subfloat{\includegraphics[width=0.16\linewidth]{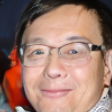}} \hfill
    \end{minipage}
    \vspace{-0.5em}
    \caption{Sample face images leaked from training data (first row) of the generative model in the \textbf{IDiff-Face (Uniform)} dataset (second row).}
    \label{fig:sample_idifface_u}
    \vspace{-4em}
\end{figure}

\begin{figure}[htbp]
    \centering
    \begin{minipage}{0.05\textwidth}
        \rotatebox{90}{Real}
    \end{minipage}%
    \begin{minipage}{0.93\textwidth}
        % First row of subfigures
        \subfloat{\includegraphics[width=0.16\linewidth]{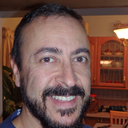}} \hfill
        \subfloat{\includegraphics[width=0.16\linewidth]{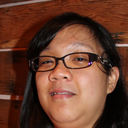}} \hfill
        \subfloat{\includegraphics[width=0.16\linewidth]{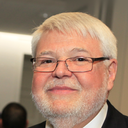}} \hfill
        \subfloat{\includegraphics[width=0.16\linewidth]{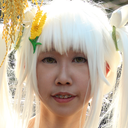}} \hfill
        \subfloat{\includegraphics[width=0.16\linewidth]{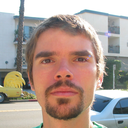}} \hfill
        \subfloat{\includegraphics[width=0.16\linewidth]{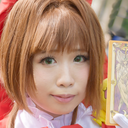}} \hfill
        \\\vspace{-9.5pt}
    \end{minipage}

    \begin{minipage}{0.05\textwidth}
        \rotatebox{90}{\hspace{5pt}Synthetic}
    \end{minipage}%
    \begin{minipage}{0.93\textwidth}
        % Second row of subfigures
        \subfloat{\includegraphics[width=0.16\linewidth]{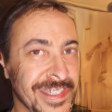}} \hfill
        \subfloat{\includegraphics[width=0.16\linewidth]{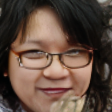}} \hfill
        \subfloat{\includegraphics[width=0.16\linewidth]{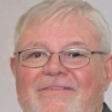}} \hfill
        \subfloat{\includegraphics[width=0.16\linewidth]{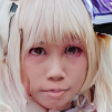}} \hfill
        \subfloat{\includegraphics[width=0.16\linewidth]{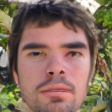}} \hfill
        \subfloat{\includegraphics[width=0.16\linewidth]{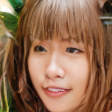}} \hfill
    \end{minipage}
    \vspace{-0.5em}
    \caption{Sample face images leaked from training data (first row) of the generative model in the \textbf{IDiff-Face (Two-stage)} dataset (second row).}
    \label{fig:sample_idifface_t}
    \vspace{-2em}
\end{figure}

\begin{figure}[htbp]
    \centering
    \begin{minipage}{0.05\textwidth}
        \rotatebox{90}{Real}
    \end{minipage}%
    \begin{minipage}{0.93\textwidth}
        % First row of subfigures
        \subfloat{\includegraphics[width=0.16\linewidth]{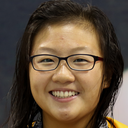}} \hfill
        \subfloat{\includegraphics[width=0.16\linewidth]{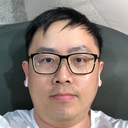}} \hfill
        \subfloat{\includegraphics[width=0.16\linewidth]{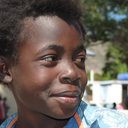}} \hfill
        \subfloat{\includegraphics[width=0.16\linewidth]{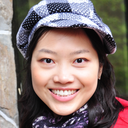}} \hfill
        \subfloat{\includegraphics[width=0.16\linewidth]{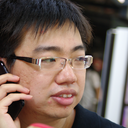}} \hfill
        \subfloat{\includegraphics[width=0.16\linewidth]{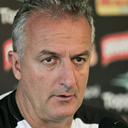}} \hfill
        \\\vspace{-9.5pt}
    \end{minipage}

    \begin{minipage}{0.05\textwidth}
        \rotatebox{90}{\hspace{5pt}Synthetic}
    \end{minipage}%
    \begin{minipage}{0.93\textwidth}
        % Second row of subfigures
        \subfloat{\includegraphics[width=0.16\linewidth]{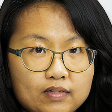}} \hfill
        \subfloat{\includegraphics[width=0.16\linewidth]{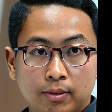}} \hfill
        \subfloat{\includegraphics[width=0.16\linewidth]{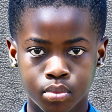}} \hfill
        \subfloat{\includegraphics[width=0.16\linewidth]{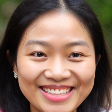}} \hfill
        \subfloat{\includegraphics[width=0.16\linewidth]{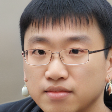}} \hfill
        \subfloat{\includegraphics[width=0.16\linewidth]{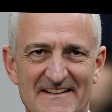}} \hfill
    \end{minipage}
    \caption{Sample face images leaked from training data (first row) of the generative model in the \textbf{GANDiffFace} dataset (second row).}
    \label{fig:sample_gandifface}
    \vspace{-2em}
\end{figure}

\begin{figure}[htbp]
    \centering
    \begin{minipage}{0.05\textwidth}
        \rotatebox{90}{Real}
    \end{minipage}%
    \begin{minipage}{0.93\textwidth}
        % First row of subfigures
        \subfloat{\includegraphics[width=0.16\linewidth]{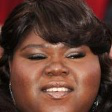}} \hfill
        \subfloat{\includegraphics[width=0.16\linewidth]{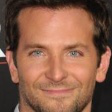}} \hfill
        \subfloat{\includegraphics[width=0.16\linewidth]{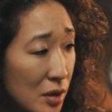}} \hfill
        \subfloat{\includegraphics[width=0.16\linewidth]{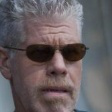}} \hfill
        \subfloat{\includegraphics[width=0.16\linewidth]{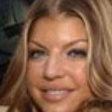}} \hfill
        \subfloat{\includegraphics[width=0.16\linewidth]{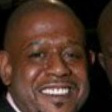}} \hfill
        \\\vspace{-9.5pt}
    \end{minipage}

    \begin{minipage}{0.05\textwidth}
        \rotatebox{90}{\hspace{5pt}Synthetic}
    \end{minipage}%
    \begin{minipage}{0.93\textwidth}
        % Second row of subfigures
        \subfloat{\includegraphics[width=0.16\linewidth]{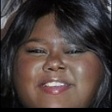}} \hfill
        \subfloat{\includegraphics[width=0.16\linewidth]{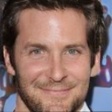}} \hfill
        \subfloat{\includegraphics[width=0.16\linewidth]{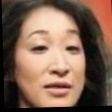}} \hfill
        \subfloat{\includegraphics[width=0.16\linewidth]{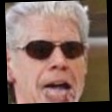}} \hfill
        \subfloat{\includegraphics[width=0.16\linewidth]{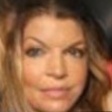}} \hfill
        \subfloat{\includegraphics[width=0.16\linewidth]{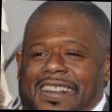}} \hfill
    \end{minipage}
    \caption{Sample face images leaked from training data (first row) of the generative model in the \textbf{IDNet} dataset (second row).}
    \label{fig:sample_idnet}
    \vspace{-2em}
\end{figure}

\begin{figure}[htbp]
    \centering
    \begin{minipage}{0.05\textwidth}
        \rotatebox{90}{Real}
    \end{minipage}%
    \begin{minipage}{0.93\textwidth}
        % First row of subfigures
        \subfloat{\includegraphics[width=0.16\linewidth]{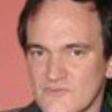}} \hfill
        \subfloat{\includegraphics[width=0.16\linewidth]{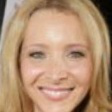}} \hfill
        \subfloat{\includegraphics[width=0.16\linewidth]{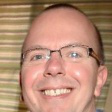}} \hfill
        \subfloat{\includegraphics[width=0.16\linewidth]{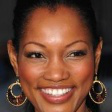}} \hfill
        \subfloat{\includegraphics[width=0.16\linewidth]{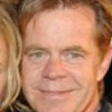}} \hfill
        \subfloat{\includegraphics[width=0.16\linewidth]{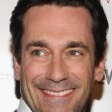}} \hfill
        \\\vspace{-9.5pt}
    \end{minipage}

    \begin{minipage}{0.05\textwidth}
        \rotatebox{90}{\hspace{5pt}Synthetic}
    \end{minipage}%
    \begin{minipage}{0.93\textwidth}
        % Second row of subfigures
        \subfloat{\includegraphics[width=0.16\linewidth]{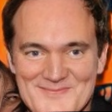}} \hfill
        \subfloat{\includegraphics[width=0.16\linewidth]{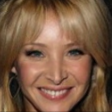}} \hfill
        \subfloat{\includegraphics[width=0.16\linewidth]{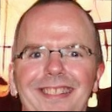}} \hfill
        \subfloat{\includegraphics[width=0.16\linewidth]{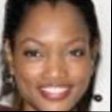}} \hfill
        \subfloat{\includegraphics[width=0.16\linewidth]{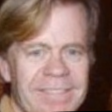}} \hfill
        \subfloat{\includegraphics[width=0.16\linewidth]{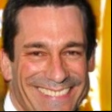}} \hfill
    \end{minipage}
    \caption{Sample face images leaked from training data (first row) of the generative model in the \textbf{SFace} dataset (second row).}
    \label{fig:sample_sface}
    \vspace{-2em}
\end{figure}

\begin{figure}[tbph]
    \centering
    \subfloat{\includegraphics[page=1,width=1\linewidth,trim=1cm 17cm 1cm 1cm,clip]{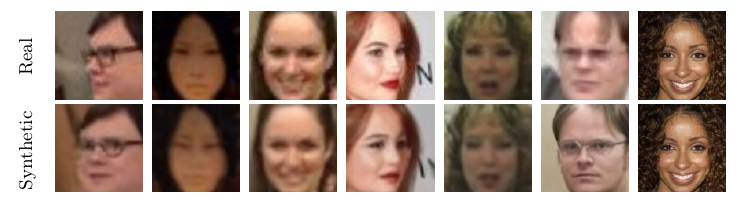}} \hfill
    \subfloat{\includegraphics[page=2,width=1\linewidth,trim=1cm 17cm 1cm 1cm,clip]{images/pdf_samples_leaked_more.pdf}} \hfill
    \subfloat{\includegraphics[page=3,width=1\linewidth,trim=1cm 17cm 1cm 1cm,clip]{images/pdf_samples_leaked_more.pdf}} \hfill
    \subfloat{\includegraphics[page=4,width=1\linewidth,trim=1cm 17cm 1cm 1cm,clip]{images/pdf_samples_leaked_more.pdf}} \hfill
    \subfloat{\includegraphics[page=5,width=1\linewidth,trim=1cm 17cm 1cm 1cm,clip]{images/pdf_samples_leaked_more.pdf}} \hfill
    \vspace{-10pt}
    \caption{Sample face images leaked from training data (CASIA-WebFace) of the generative model in the \textbf{DCFace} dataset.}
    \label{fig:sample_dcface_more}
\end{figure}